\documentclass[conference]{IEEEtran}
\IEEEoverridecommandlockouts
\usepackage{cite}
\usepackage{amsfonts,bm}
\usepackage{algorithmic}
\usepackage{graphicx}
\usepackage{textcomp}
\usepackage{xcolor}
\usepackage{amsmath}
\usepackage{amssymb}
\usepackage{makecell}
\usepackage{booktabs}
\usepackage{multirow}
\usepackage{siunitx}
\usepackage{url}
\usepackage{diagbox}

\usepackage{colortbl}

\usepackage{isomath}

\usepackage{dsfont}
\usepackage{etoolbox}

\usepackage{mathtools}
\DeclarePairedDelimiter\abs{\lvert}{\rvert}%
\makeatletter
\let\oldabs\abs
\def\abs{\@ifstar{\oldabs}{\oldabs*}}

\def\A{{\mathbf A}}

\def\Y{{\mathbf Y}}
\def\y{{\mathbf y}}

\def\BibTeX{{\rm B\kern-.05em{\sc i\kern-.025em b}\kern-.08em
    T\kern-.1667em\lower.7ex\hbox{E}\kern-.125emX}}

\begin{document}

\title{Evaluating Short-Term Forecasting of Multiple Time Series in IoT Environments\\
}


\author{Christos Tzagkarakis\textsuperscript{1}, Pavlos Charalampidis\textsuperscript{1}, Stylianos Roubakis\textsuperscript{1},\\ Alexandros Fragkiadakis\textsuperscript{1} and Sotiris Ioannidis\textsuperscript{2,1}\\
\IEEEauthorblockA{\textsuperscript{1}\textit{Foundation for Research and Technology-Hellas, Institute of Computer Science, Heraklion, Greece}\\
\textsuperscript{2}\textit{Technical University of Crete, Chania, Greece}\\
E-mails: \{tzagarak, pcharala, roub, alfrag\}@ics.forth.gr, sotiris@ece.tuc.gr}
}

\maketitle

\begin{abstract}

Modern Internet of Things (IoT) environments are monitored via a large
number of IoT enabled sensing devices, with the data acquisition and
processing infrastructure setting restrictions in terms of computational
power and energy resources. To alleviate this issue, sensors are often
configured to operate at relatively low sampling frequencies, yielding a
reduced set of observations. Nevertheless, this can hamper dramatically
subsequent decision-making, such as forecasting.
To address this problem, in this work we evaluate short-term forecasting in highly underdetermined cases, i.e., the number of sensor streams is much higher than the number
of observations. Several statistical, machine learning and neural network-based models are thoroughly examined with respect to the resulting forecasting accuracy on five different real-world datasets. The focus is given on a unified experimental protocol especially designed for short-term prediction of multiple time series at the IoT edge. The proposed framework can be considered as an important step towards establishing a solid forecasting strategy in resource constrained IoT applications.

\end{abstract}


\section{Introduction}
With the advent of the Internet of Things (IoT) era, the concept of Internet networking
has been shifted into a ubiquitous interconnection between users, data and smart devices or sensors in a seamless way~\cite{6714496},~\cite{6774858}.
A wide range of applications has emerged at the forefront of IoT technology spanning from smart cities~\cite{6740844},~\cite{8030479}, smart homes~\cite{RISTESKASTOJKOSKA20171454},~\cite{6516934} and wearables~\cite{6844949},~\cite{6780609} to energy management~\cite{SHROUF2015235}, 
predictive maintenance~\cite{7520653}, automotive driving~\cite{6709775}, etc. The modern IoT systems are typically decomposed into three levels, i.e., highly heterogeneous data
are captured from the surrounding environment via several IoT devices/sensors at the edge level, which then can be transmitted through the fog layer up to the cloud for further storage and processing. However, the rapidly growing use and realization of IoT technology comes at the cost
of resolving major technical and business impediments as
reflected in dynamicity, scalability and heterogeneity~\cite{7322178},~\cite{LEE2015431}.

Specifically, a dynamically adaptive behaviour is followed
at the IoT infrastructure, at the IoT applications and at the
IoT devices, and thus it is imperative to promote a (semi)-automatic behaviour within all IoT layers. This gives rise to the
pursuit of high scalability properties from the network layers
as well as from the IoT infrastructure.
In addition, enhanced
heterogeneous behaviour as a result of the extensive use and
interconnection of a large volume of diverse IoT devices
should be addressed through the concept of efficient semantic interoperability within IoT applications and platforms.
These IoT inter-layer characteristics could promote a direct data analysis at the IoT edge as a very important factor within IoT systems.

Typically, a large number of sensing devices is used to monitor  IoT environments. The data acquisition and processing modules impose restrictions in terms of computational power and energy resources at the edge level. These constraints can be handled by configuring the sensors to operate at relatively low sampling frequencies, yielding a reduced set of observations, which can be further analyzed by adopting lightweight algorithmic procedures. This motivates us to investigate the short-term forecasting of multiple sensor streams at the IoT edge. Specifically, we assume that a set of sensing devices is used to collect time series data from the IoT environment, 
where the number of time series is much higher than the number of observations per sensor. Instead of predicting the future values of each sensor stream separately, each sensor's collected time series can be transmitted from the edge to an aggregation mechanism such as an edge gateway. As such, an overall (global) forecasting model can be computed based on the aggregated multiple time series, considering in this way the inherent relationships between the acquired time series.

\noindent \textbf{Related work.} Recent works have dealt with the multiple time series forecasting task by focusing on IoT-related data. A block Hankel tensor-based autoregressive integrated moving average (BHT-ARIMA) method is described in~\cite{BHTARIMA} that exploits the intrinsic correlations among multiple time series. The authors in~\cite{9075431} introduce a long short-term memory multiseasonal network (LSTM-MSNet), i.e., a three-layered forecasting framework using LSTMs that accounts for the multiple seasonal periods present in time series. Short-term electricity load forecasting is examined in~\cite{Dudek2015}, where random forest (RF) models are used to forecast time series with multiple seasonal variations. In~\cite{Monashforecasting2021} a comprehensive time series forecasting benchmarking archive is proposed that contains twenty-five publicly available time series datasets from various domains, with different characteristics in terms of series lengths, time resolution,  and inclusion of missing values.

\noindent \textbf{Contribution.} The contribution of the current paper is twofold. Firstly,  we establish a solid multiple time series forecasting protocol for single-step prediction, especially targeted in IoT use cases. Secondly, a set of five different real-world IoT datasets is thoroughly examined by using off-the-shelf statistical, machine learning and neural network methods as well as scale-free and percentage error-based accuracy metrics.





The rest of the paper is organized as follows: Section~\ref{sec:problem_formulation} overviews the problem formulation. The proposed multiple time series prediction protocol is described in Section~\ref{sec:forecasting_procedure}, while Section~\ref{sec:experimental_eval}
demonstrates the experimental evaluation in light of the forecasting accuracy on five real-world datasets. Finally, Section~\ref{sec:conclusions} summarizes the main results and gives directions for further extensions.


\section{Problem Formulation}\label{sec:problem_formulation}
Let us assume that a collection of $N$ equal length time series can be written in matrix form 
\begin{equation}
    \Y= \begin{bmatrix}
    \y_1 \\
    \vdots \\
    \y_l \\
    \vdots \\
    \y_L
    \end{bmatrix}=
    \begin{bmatrix}
    y_1^1 & y_1^2 & \ldots & y_1^N  \\
    \vdots & \vdots & & \vdots \\
    y_l^1 & y_l^2 & \ldots & y_l^N \\
    \vdots & \vdots & & \vdots \\
    y_L^1 & y_L^2 & \ldots & y_L^N
    \end{bmatrix}\in\mathbb{R}^{L\times N},
    \label{eq:matrix_Y_definition}
\end{equation}
where each row $\y_l=[y_l^1 \ldots y_l^N]\in\mathbb{R}^{1\times N}$, $l\in\{1,\ldots,L\}$, represents the $l$-th observation of all time series and each column $\y^n=[y_1^n \ldots y_L^n]^T\in\mathbb{R}^{N\times 1}$, $n\in\{1,\ldots,N\}$, denotes the time series captured by the $n$-th sensor. Our main goal is to compute a forecasting model $\mathcal{M}$, which uses historical observations of the multiple time series $\Y$ to predict the future values 
\begin{equation}
    \Y^{(f)}= \begin{bmatrix}
    \y_{L+1} \\
    \vdots \\
    \y_{L+h}
    \end{bmatrix}=
    \begin{bmatrix}
    y_{L+1}^1 & y_{L+1}^2 & \ldots & y_{L+1}^N  \\
    \vdots & \vdots & & \vdots \\
    y_{L+h}^1 & y_{L+h}^2 & \ldots & y_{L+h}^N
    \end{bmatrix}\in\mathbb{R}^{h\times N}
    \label{eq:matrix_pred_definition}
\end{equation}
for a given forecasting horizon $h$.
In this work, we assume a single-step forecasting horizon $h=1$, and thus the matrix~(\ref{eq:matrix_pred_definition}) collapses into a vector 
\begin{equation}
\y^{(f)}=[y_{L+1}^1\, y_{L+1}^2 \,\ldots\, y_{L+1}^N]\in\mathbb{R}^{1\times N}.
\label{eq:original_samples}
\end{equation}
A prediction model $\mathcal{M}$ can then be defined as
\begin{equation}
    \hat{\y}^{(f)}=\mathcal{M}(\Y,\boldsymbol\theta),
    \label{eq:model_M}
\end{equation}
where
\begin{equation}
\hat{\y}^{(f)}=[\hat{y}_{L+1}^1\, \hat{y}_{L+1}^2\, \ldots\, \hat{y}_{L+1}^N]\in\mathbb{R}^{1\times N}
\label{eq:pred_samples}
\end{equation}
denotes the predicted samples within the single-step forecasting horizon, and $\boldsymbol\theta$ are the model parameters. It is obvious that the smaller the difference between the actual $\y^{(f)}$ and the predicted $\hat{\y}^{(f)}$ samples, the more accurate the future samples prediction is.

\subsection{Forecasting Models}\label{sec:pm}
The forecasting model $\mathcal{M}$ can be estimated through various approaches grouped into three main categories, namely the statistical, the machine learning and the neural network methods. 
BHT-ARIMA~\cite{BHTARIMA} is the statistical  model used in the current work,
while support vector regression (SVR)~\cite{BAO2014482} and RF model~\cite{Dudek2015} correspond to the adopted machine learning approaches. The used neural network-based methods correspond to long short-term memory (LSTM), bidirectional LSTM (Bi-LSTM) and convolutional neural networks (CNN)~\cite{Benidis2020},~\cite{9005997} as well as to echo state network (ESN)~\cite{GALLICCHIO201833}. Each forecasting model's input data should be transformed into the appropriate shape for training, hyper-parameter tuning and testing purposes.  Next, more details on the data shaping process and the overall forecasting protocol are provided.

\section{Forecasting Protocol}\label{sec:forecasting_procedure}
A specific sequence of steps must be followed to estimate the future samples  $\y^{(f)}$ through the forecasting model $\mathcal{M}$. As a first step, it is important to perform data scaling, since in real-world IoT applications, the collected time series contain observations with different value ranges.
The min-max scaler
\begin{equation}
    \y^n_{sc} = \frac{\y^n-\y^n_{min}}{\y^n_{max}-\y^n_{min}},\,\,{n=1,\ldots,N}
    \label{eq:MinMax_scale}
\end{equation}
is used for data normalization, with $\y^n_{sc}$ representing the scaled observations of the $n$-th time series, while $\y^n_{min}$ and $\y^n_{max}$ are the minimum and maximum values of the $n$-th time series, respectively. The scaled version of the multiple time series~(\ref{eq:matrix_Y_definition}) can be written as $\Y_{sc}\in\mathbb{R}^{L\times N}$.

The second step involves data training/validation of each forecasting model mentioned in Section~\ref{sec:pm}. To properly train and validate each forecasting model based on the available data, we need to follow a rolling window-based cross-validation procedure towards identifying the best hyper-parameter configuration per model. In the following subsections, we describe the different types of cross-validation methods per forecasting model category. Before proceeding, let us define the notations: $^{[i]}\A$ denotes the $i$-th row of matrix $\A$, $^{[i:j]}\A$ is the submatrix of $\A$ from row $i$ to row $j$ and $\hat{\Y}_{sc}$ is the predicted counterpart of $\Y_{sc}$.


\subsection{Rolling Window: Matrix Case}\label{sec:rolling_window1}
In the case of SVR and RF machine learning models, it is necessary to perform hyper-parameter tuning by using the pair of training data submatrices $\left({^{[1:L_{tr}-1]}}\Y_{sc},\, {^{[L_{tr}]}}\Y_{sc}\right)$, where $L_{tr}<L$ defines the amount of training data in each cross-validation iteration. 
In particular, each machine learning model is trained on  this pair of data submatrices, where the trained (for a specific hyper-parameter configuration) model's forecasting performance is then evaluated on the first validation fold ${^{[L_{tr}+1]}}\Y_{sc}$, obtaining the validation error of the form
\begin{equation}
e_{v}^1 = \mathcal{J}\Big({^{[L_{tr}+1]}}\Y_{sc},{^{[L_{tr}+1]}}\hat{\Y_{sc}}\Big),
\label{eq:first_val_error}
\end{equation}
where $\mathcal{J}$ is the function which determines the forecasting accuracy between the original multiple time samples and the predicted ones.
Next, the second training data pair $\left({^{[2:L_{tr}]}}\Y_{sc},\, {^{[L_{tr}+1]}}\Y_{sc}\right)$ is composed by sliding the training window one step forward, yielding the second validation error 
\begin{equation}
e_{v}^2=\mathcal{J}\Big({^{[L_{tr}+2]}}\Y_{sc},{^{[L_{tr}+2]}}\hat{\Y_{sc}}\Big)
\label{eq:second_val_error}
\end{equation}
 for the same hyper-parameter configuration.
We keep moving the window forward one step at a time until the computation of the last validation error
\begin{equation}
e_{v}^{L_v}=\mathcal{J}\Big({^{[L_{tr}+L_v]}}\Y_{sc}, {^{[L_{tr}+L_v]}}\hat{\Y_{sc}}\Big)
\label{eq:last_val_error}
\end{equation}
based on the trained model fed with the data $\left({^{[L_{tr}-1:L_{tr}+L_{v}-2]}}\Y_{sc},\, {^{[L_{tr}+L_{v}-1]}}\Y_{sc}\right)$, where $L_v=L-L_{tr}$ denotes the amount of validation data. The best hyper-parameter configuration corresponds to the minimum mean validation error 
\begin{equation}
    \overline{e}_{v}=\frac{1}{L_v}\sum_{k=1}^{L_{v}}e_{v}^{k}.
    \label{eq:mean_val_error}
\end{equation}

For BHT-ARIMA, we follow a similar hyper-parameter tuning concept by providing ${^{[1:L_{tr}]}}\Y_{sc}$ as training data input to the model during the first cross-validation iteration, and computing the first validation error~(\ref{eq:first_val_error}). The training window is then shifted forward by a single step to obtain the second training data input $^{[2:L_{tr}+1]}\Y_{sc}$, and thus providing the second validation error~(\ref{eq:second_val_error}). The rolling window process is repeated until the last training segment ${^{[L_{tr}-1:L_{tr}+L_{v}-1]}}\Y_{sc}$, with the last validation error computed as in~(\ref{eq:last_val_error}). The optimal hyper-parameter combination corresponds to the minimum mean validation error~(\ref{eq:mean_val_error}).

\subsection{Rolling Window: Multi-dimensional Matrix Case}\label{sec:rolling_window2}
In the case of LSTM, Bi-LSTM and CNN neural network models, a slightly different rolling window-based cross-validation methodology is followed.
Specifically, an additional hyper-parameter $S<L_{tr}$ is introduced, which determines the rolling window length.
Next, a pair of data $\left({^{[1:S]}}\Y_{sc},\,{^{[S+1]}}\Y_{sc}\right)$ is formed. The rolling window of length $S$ is then shifted forward by a single step composing the second pair $\left(^{[2:S+1]}\Y_{sc},\,^{[S+2]}\Y_{sc}\right)$ of data. This rolling window scheme is repeated until the assembly of the last pair $\left(^{[L_{tr}-S:L_{tr}-1]}\Y_{sc},\,^{[L_{tr}]}\Y_{sc}\right)$. To properly feed the LSTM, Bi-LSTM and CNN model with the training input data, we need to concatenate these pairs as follows
\begin{equation}
    \begin{split}
    \left( \left[{^{[1:S]}}\Y_{sc}\,,\,{^{[2:S+1]}}\Y_{sc}\,,\ldots,\,{^{[L_{tr}-S:L_{tr}-1]}}\Y_{sc}\right], \right. \\
    \left. \left[{^{[S+1]}}\Y_{sc}\,,\,{^{[S+2]}}\Y_{sc}\,,\ldots,\,{^{[L_{tr}]}}\Y_{sc}\right]  \right).
    \end{split}
\end{equation}
The trained neural network model (for a particular hyper-parameter combination) is fed with $^{[L_{tr}-S+1:L_{tr}]}\Y_{sc}$ to compute the first validation error~(\ref{eq:first_val_error}).
The second validation error~(\ref{eq:second_val_error}) is estimated based on the data $^{[L_{tr}-S+2:L_{tr}+1]}\Y_{sc}$, while the last validation error~(\ref{eq:last_val_error})  is obtained given the data ${^{[L_{tr}-S+L_{v}:L_{tr}+L_{v}-1]}}\Y_{sc}$, and
the optimal hyper-parameter configuration corresponds to the minimum mean validation error~(\ref{eq:mean_val_error}).

\subsection{Rolling Window: Matrix List Case}\label{sec:rolling_window3}
Here, we overview the rolling window-based cross-validation methodology in the case of ESN model. In particular, let us introduce an extra hyper-parameter $S<L_{tr}$ as in the previous subsection. The following training pairs
\begin{equation}
    \begin{split}
    \left( \left({^{[1:S]}}\Y_{sc},\,{^{[2:S+1]}}\Y_{sc}\right),\left({^{[2:S+1]}}\Y_{sc},\,{^{[3:S+2]}}\Y_{sc}\right),\ldots, \right. \\
    \left. \left({^{[L_{tr}-S:L_{tr}-1]}}\Y_{sc},\,{^{[L_{tr}-S+1:L_{tr}]}}\Y_{sc}\right)  \right)
    \end{split}
\end{equation}
are then composed and used as input to train the ESN model (given a specified hyper-parameter combination).
After the ESN training process, the pair of data $\left(^{[L_{tr}-S:L_{tr}-1]}\Y_{sc},\,^{[L_{tr}-S+1:L_{tr}]}\Y_{sc}\right)$ is fed as input to the trained ESN model in order to compute the first validation error~(\ref{eq:first_val_error}). The second validation error~(\ref{eq:second_val_error}) is estimated based on the pair of data
$\left(^{[L_{tr}-S+1:L_{tr}]}\Y_{sc},\,^{[L_{tr}-S+2:L_{tr}+1]}\Y_{sc}\right)$. This rolling window process is repeated until the last training data input $\left({^{[L_{tr}+L_v-S-1:L_{tr}+L_v-2]}}\Y_{sc},\,{^{[L_{tr}+L_v-S:L_{tr}+L_v-1]}}\Y_{sc}\right)$ provides the last validation error~(\ref{eq:last_val_error}). As in the two rolling window schemes above, the minimum mean validation error~(\ref{eq:mean_val_error}) provides the optimal hyper-parameter configuration.

\section{Experimental Evaluation}\label{sec:experimental_eval}
In this section, we evaluate the short-term multiple time series forecasting models on real-world time series datasets based on the protocol described in Section~\ref{sec:forecasting_procedure}.

\subsection{Datasets Description}\label{sec:dataset_info}
We use five publicly available multiple time series datasets that correspond to various practical IoT applications. 
Specifically, the \textit{Energy Consumption Fraunhofer} dataset\footnote{\url{https://fordatis.fraunhofer.de/handle/fordatis/215}} contains the measured energy consumption, with an hourly time resolution, of $499$ customers (each customer is assigned to one of the $68$ customer profiles such as private households, shops, bakeries) in Spain. The energy consumption is measured in kilowatt hour (kWh). Each time series contains the consumption values within the measurement period of Jan. 1 to Dec. 31, 2019.
In the current experimental framework, the mean daily energy consumption of each customer's household is used, and thus the final dataset consists of $314$ time series (customers) and 365 (daily) measurements per time series.

The  \textit{Electricity Load Diagrams} dataset\footnote{\url{https://archive.ics.uci.edu/ml/datasets/ElectricityLoadDiagrams20112014}} corresponds to electricity measurements across $370$ households in Portugal from $2011$ to $2014$, with a $15$ minutes time resolution.
Here, we use the total amount of energy consumption per customer in a daily basis during the period Jan. $1$, $2012$ to Dec. $31$, $2014$, providing an overall dataset of size $324$ time series by $1096$ measurements.

The \textit{Guangzhou Traffic} dataset\footnote{\url{https://zenodo.org/record/1205229\#.YguzWN9BxaR}} is an urban traffic speed dataset collected in Guangzhou, China,  which consists of $214$ anonymous road segments within the period Aug. $1$ to Sep. $30$, $2016$ at $10$ minutes interval.
Since the original dataset contains missing values, the road segments containing non-zero values are only kept. During the experimental evaluation, we use the mean hourly traffic speed per road segment, and thus the final dataset's size is $206$ time series (road segments) by $1464$ measurements.

The \textit{San Francisco Traffic} dataset\footnote{\url{https://zenodo.org/record/4656135\#.Ygu-o99BxaQ}} 
is a collection of hourly time series,
representing the traffic occupancy rate of different car
lanes of San Francisco bay area freeways between 2015 and 2016. The size of the used dataset is $862$ time series by $104$ measurements.

The \textit{London Smart Meters} dataset\footnote{\url{https://zenodo.org/record/4656091\#.YgvTLd9BxaS}} consists of $5560$ half hourly time series that represent the energy consumption readings of London households in kWh from Nov. 2011 to Feb. 2014. Here, we select $504$ households with energy consumption during Oct. 23, $2012$ until May 18, $2013$ leading to $9983$ measurements per time series.
The overall datasets information is presented in Table~\ref{tbl:DatasetsInfo}.
\begin{table}[t]
    \caption{Datasets information}\label{tbl:DatasetsInfo}
	\centering
	\resizebox{\columnwidth}{!}{
	\begin{tabular}{l c c c}
	 \hline
	 Dataset & \makecell{Number of\\ time series} & \makecell{Number of\\ samples per\\ time series} & \makecell{Time\\ resolution} \\
	 \hline\hline
	Energy Consumption Fraunhofer (D1) & 314 & 365 & daily\\
	Electricity Load Diagrams (D2) & 320 & 1096 & daily \\
	Guangzhou Traffic (D3) & 206 & 1464 & hourly\\
	San Francisco Traffic (D4) & 862 & 104 & weekly \\
	London Smart Meters (D5) & 504 & 9983 & half hourly\\
	 \hline
	\end{tabular}
	}
\end{table}


\begin{table*}[!t]
\setlength{\tabcolsep}{3.pt}
\begin{center}
\caption{Forecasting results}
\label{tbl:forecasting_results}
\begin{tabular}{cc|cc||cc||cc||cc||cc}
\hline
& \textsc{Dataset} & \multicolumn{2}{c}{\textsc{D1}} & \multicolumn{2}{c}{\textsc{D2}} & \multicolumn{2}{c}{\textsc{D3}} & \multicolumn{2}{c}{\textsc{D4}} & \multicolumn{2}{c}{\textsc{D5}}\\ \cline{3-4} \cline{5-6} \cline{7-8} \cline{9-10} \cline{11-12}\hline
 & \textbf{\# Sim. data} & \textbf{40} & \textbf{90} & \textbf{40} & \textbf{90} & \textbf{40} & \textbf{90} & \textbf{40} & \textbf{90} & \textbf{40} & \textbf{90}\\ \hline\hline
 

& sMAPE & 0.3369 & 0.3554 & 0.0969 & 0.0819 & 0.0797 & 0.1102 & \cellcolor{cyan!20}{0.0742} & 0.1022 &	0.5209 & 0.4683\\
\textbf{BHT-ARIMA} & MAAPE & \cellcolor{green!20}{0.2888} & \cellcolor{green!20}{0.3035} & 0.0986 & 0.0756 & 0.0765 & 0.1050 & \cellcolor{green!20}{0.0724} & 0.1025 & \cellcolor{green!20}{0.4298} & \cellcolor{green!20}{0.3988}\\
& MASE & \cellcolor{gray!20}{0.1806} & \cellcolor{gray!20}{0.1905} & 0.0503 & \cellcolor{gray!20}{0.0247} & 0.3169 & 0.3569 & \cellcolor{gray!20}{0.1877} & 0.2576 & 0.7366 & 0.3829\\\hline

& sMAPE & 0.3683 & 0.3816 & \cellcolor{cyan!20}{0.0931} & 0.0765 & 0.0655 & 0.0930 & 0.0795 & \cellcolor{cyan!20}{0.0946} & 0.4705 & 0.5047\\
\textbf{SVR} & MAAPE & 0.3275 & 0.3424 & \cellcolor{green!20}{0.0933} & 0.0732 & 0.0666 & 0.0935 & 0.0766 & \cellcolor{green!20}{0.0952} & 0.4383 & 0.4898\\
& MASE & 0.1867 & 0.1945 & \cellcolor{gray!20}{0.0421} & 0.0250 & 0.2697 & 0.3190 & 0.1978 & \cellcolor{gray!20}{0.2267} & \cellcolor{gray!20}{0.3878} & 0.4271\\\hline

& sMAPE & \cellcolor{cyan!20}{0.3200} & \cellcolor{cyan!20}{0.3417} & 0.1026 & 0.0813 & 0.0601 & 0.0921 & 0.0814 & 0.1076 & \cellcolor{cyan!20}{0.4649} & \cellcolor{cyan!20}{0.4353}\\
\textbf{RF} & MAAPE & 0.2907 & 0.3086 & 0.1033 & 0.0782 & 0.0615 & 0.0923 & 0.0788 & 0.1068 & 0.4461 & 0.4281\\
& MASE & 0.1892 & 0.1927 & 0.0459 & 0.0248 & 0.2324 & 0.2978 & 0.2037 & 0.2617 & 0.3935 & \cellcolor{gray!20}{0.3673}\\\hline

& sMAPE & 0.5551 & 0.6310 & 0.1668 & 0.2151 & 0.4762 & 0.5625 & 0.1771 & 0.2652 & 0.9009 & 0.8736\\
\textbf{LSTM} & MAAPE & 0.3672 & 0.4038 & 0.1373 & 0.1597 & 0.3281 & 0.3675 & 0.1488 & 0.2204 & 0.4959 & 0.4842\\
& MASE & 0.3111 & 0.3557 & 0.0849 & 0.0885 & 1.8338 & 2.0560 & 0.4068 & 0.5719 & 0.6320 & 0.6127\\\hline

& sMAPE & 0.5557 & 0.6299 & 0.1667 & 0.2138 & 0.4761 & 0.5515 & 0.1770 & 0.3246 & 0.9017 & 0.8734\\
\textbf{Bi-LSTM} & MAAPE & 0.3682 & 0.4048 & 0.1373 & 0.1590 & 0.3280 & 0.3623 & 0.1487 & 0.2577 & 0.4963 & 0.4843\\
& MASE & 0.3112 & 0.3558 & 0.0849 & 0.0883 & 1.8338 & 2.0253 & 0.4067 & 0.7376 & 0.6321 & 0.6124\\\hline

& sMAPE & 0.5555 & 0.6314 & 0.1668 & 0.2150 & 0.4738 & 0.5619 & 0.1771 & 0.2639 & 0.9010 & 0.8732\\
\textbf{CNN} & MAAPE & 0.3680 & 0.4046 & 0.1374 & 0.1596 & 0.3266 & 0.3672 & 0.1488 & 0.2191 & 0.4958 & 0.4845\\
& MASE & 0.3113 & 0.3562 & 0.0849 & 0.0883 & 1.8235 & 2.0542 & 0.4069 & 0.5678 & 0.6322 & 0.6124\\\hline

& sMAPE & 0.3510 & 0.4262 & 0.1021 & \cellcolor{cyan!20}{0.0749} & \cellcolor{cyan!20}{0.0588} & \cellcolor{cyan!20}{0.0825} & 0.0825 & 0.1216 & 0.5154 & 0.6125\\
\textbf{ESN} & MAAPE & 0.3084 & 0.3657 & 0.1030 & \cellcolor{green!20}{0.0719} & \cellcolor{green!20}{0.0601} & \cellcolor{green!20}{0.0830} & 0.0801 & 0.1206 & 0.4562 & 0.5251\\
& MASE & 0.1897 & 0.2240 & 0.0474 & 0.0256 & \cellcolor{gray!20}{0.2322} & \cellcolor{gray!20}{0.2954} & 0.2054 & 0.2714 & 0.4239 & 0.5068\\\hline
\end{tabular} 
\end{center}
\end{table*}

\subsection{Performance Metrics}\label{sec:performance_metrics}
We employ percentage error-based and scale-free error metrics to assess the accuracy of the forecasting models. Three error metrics are used in the context of multiple time series short-term forecasting within a prediction horizon $h=1$, namely the symmetric mean absolute percentage error (sMAPE), the recently proposed mean arctangent absolute percentage error (MAAPE)~\cite{KIM2016669} and the mean absolute scaled error (MASE) given by
\begin{equation}
    \mathrm{sMAPE}\left(\y^{(f)},\hat{\y}^{(f)}\right)=\frac{1}{N}\sum_{n=1}^N 2\frac{\abs{y_{L+1}^n-\hat{y}_{L+1}^n}}{\abs{y_{L+1}^n}+\abs{\hat{y}_{L+1}^n}}
    \label{eq:smape}
\end{equation}
\begin{equation}
    \mathrm{MAAPE}\left(\y^{(f)},\hat{\y}^{(f)}\right)=\frac{1}{N}\sum_{n=1}^N\arctan\left(\abs{\frac{y_{L+1}^n-\hat{y}_{L+1}^n}{y_{L+1}^n}}\right)
    \label{eq:maape}
\end{equation}
\begin{equation}
    \mathrm{MASE}\left(\y^{(f)},\hat{\y}^{(f)}\right)=\frac{1}{N}\sum_{n=1}^N\frac{\abs{y_{L+1}^n-\hat{y}_{L+1}^n}}{\frac{1}{L-1}\sum\limits_{l=2}^L\abs{y_l^n-\hat{y}_{l}^n}},
    \label{eq:mase}
\end{equation}
where $\y^{(f)}$ and $\hat{\y}^{(f)}$ is the original and predicted samples vector, as defined in~(\ref{eq:original_samples}) and~(\ref{eq:pred_samples}), respectively.
The sMAPE is used as the function $\mathcal{J}$ during the cross-validation process, while sMAPE, MAAPE and MASE are used to evaluate the forecasting generalization (out-of-sample) performance of each prediction model.

\subsection{Hyper-parameter Tuning}\label{sec:tuning}
Based on the forecasting protocol mentioned in Section~\ref{sec:forecasting_procedure}, each dataset is split into a training (80\%) and a validation (20\%) partition.
For each prediction model, a different rolling window-based cross-validation approach is followed to identify the best hyper-parameter combination, as discussed in Section~\ref{sec:rolling_window1}-~\ref{sec:rolling_window3}. Hyper-parameter grid search is performed for each model per dataset. Due to lack of space, we provide a detailed description of the  hyper-parameter grids per model in~{\url{https://github.com/pcharala/multiple-timeseries-forecasting}}.

\subsection{Results}\label{sec:results}
In this section, we compare the forecasting performance of each prediction model $\mathcal{M}$ with the performance obtained by predicting the future single-step $\y^{(f)}$ based on the scaled multiple time series $\Y_{sc}$. For this purpose, we perform simulations on five real-world datasets described in Section~\ref{sec:dataset_info}. We investigate the forecasting accuracy of each prediction model for a varying length $L$ of $\Y_{sc}$, i.e.,  the accuracy for a total length $L=40$ and $L=90$ is examined for all datasets.
The single-step forecasting results are averaged over fifteen Monte Carlo iterations, where during each Monte Carlo run, a different part of the time series dataset is examined. The average error metric per dataset is reported to show the overall mean errors for each prediction model.

Table~\ref{tbl:forecasting_results} shows the forecasting results measured in terms of sMAPE, MAAPE and MASE. The datasets names D1 (Energy Consumption Fraunhofer), D2 (Electricity Load Diagrams), D3 (Guangzhou Traffic), D4 (San Francisco Traffic), D5 (London Smart Meters) and the respective experimental window configurations are depicted in the first two rows of the table. The best sMAPE, MAAPE and MASE value is depicted in blue, green and gray colour, respectively. As it can be seen, the statistical (BHT-ARIMA) and machine learning (SVR, RF) models achieve in general a better forecasting performance against the neural network models (LSTM, Bi-LSTM, CNN, ESN). This behaviour has been observed in other works (e.g. in~\cite{BHTARIMA}) and can be attributed to the fact that neural networks need more data to achieve a higher performance. Our analysis code and data can be found on GitHub:\\\url{https://github.com/pcharala/multiple-timeseries-forecasting}.

\section{Conclusions and future work}\label{sec:conclusions}
In this paper, we address
the problem of short-term forecasting of multiple time series in IoT environments, when
the number of time series is much higher than the number of observations. Seven different prediction models
are thoroughly examined on five real-world IoT datasets, applying a unified experimental protocol especially designed for short-term prediction of multiple time series at the IoT edge. It is experimentally shown that the statistical and machine learning models are generally outperforming the neural network-based methods. As a future work, we intend to conduct extensive simulations using a larger number of IoT datasets and lightweight multiple time-series prediction models as well as investigating the training/validation performance of additional rolling window-based cross-validation techniques.

\section*{Acknowledgment}
This work has received funding from the European Union's Horizon 2020 research and innovation programme under grant agreement No 957337 (project MARVEL) and the Operational Program Competitiveness, Entrepreneurship and Innovation, under the call RESEARCH–CREATE–INNOVATE (project code: T1EDK-00070).


\bibliographystyle{IEEEtran}
\bibliography{bibliography}

\end{document}